%% file: main.tex
\documentclass[conference]{IEEEtran}  

\ifCLASSOPTIONcompsoc
  \usepackage[nocompress]{cite}
\else
  \usepackage{cite}
\fi

\usepackage{xcolor}
\usepackage{graphicx}
\usepackage{multirow}
\usepackage{rotating}
\usepackage{subcaption}
\newcommand{\STAB}[1]{\begin{tabular}{@{}c@{}}#1\end{tabular}}

\begin{document}

\title{Towards Explaining Satellite Based Poverty Predictions with Convolutional Neural Networks}

\author{\IEEEauthorblockN{
            Hamid Sarmadi\IEEEauthorrefmark{1},
            Thorsteinn R\"{o}gnvaldsson\IEEEauthorrefmark{1},
            Nils Roger Carlsson\IEEEauthorrefmark{1},
            Mattias Ohlsson\IEEEauthorrefmark{1}\IEEEauthorrefmark{3},
            Ibrahim Wahab\IEEEauthorrefmark{2},
            Ola Hall\IEEEauthorrefmark{2}%
        }
        \IEEEauthorblockA{
            \IEEEauthorrefmark{1}Center for Applied Intelligent Systems Research (CAISR), 
            Halmstad University, Sweden\\
            \{hamid.sarmadi, thorsteinn.rognvaldsson, nils\_roger.carlsson, mattias.ohlsson\}@hh.se\\
            \IEEEauthorrefmark{2}Department of Human Geography, Lund University, Sweden\\
            \{ibrahim.wahab, ola.hall\}@keg.lu.se\\
            \IEEEauthorrefmark{3}Centre for Environmental and Climate Science, \\ Lund University, Sweden
        }
}

%
%

\IEEEtitleabstractindextext{%
\begin{abstract}
\input{Abstract}
\end{abstract}

}

\IEEEoverridecommandlockouts
 \IEEEpubid{\makebox[\columnwidth]{979-8-3503-4503-2/23/\$31.00~\copyright2023 IEEE \hfill}
 \hspace{\columnsep}\makebox[\columnwidth]{ }}
\maketitle
\IEEEpubidadjcol

\IEEEdisplaynontitleabstractindextext

%
\IEEEpeerreviewmaketitle

\section{Introduction}
\input{Introduction}

\section{Previous work}
\input{Background}

\section{Data}
\input{Data}

\section{Methods}
\input{Methods}

\section{Results}
\input{Results}

\section{Discussion}
\input{Discussion}

\section{Conclusion}
\input{Conclusion}
\bibliographystyle{IEEEtran}
\bibliography{IEEEabrv,xai}
\end{document}

%% file: Abstract.tex
Deep convolutional neural networks (CNNs) have been shown to predict poverty and development indicators from satellite images with surprising accuracy.
This paper presents a first attempt at analyzing the CNNs responses in detail and explaining the basis for the predictions. The CNN model, while trained on relatively low resolution day- and night-time satellite images, is able to outperform human subjects who look at high-resolution images in ranking the Wealth Index categories. Multiple explainability experiments performed on the model indicate the importance of the sizes of the objects, pixel colors in the image, and provide a visualization of the importance of different structures in input images. A visualization is also provided of type images that maximize the network prediction of Wealth Index, which provides clues on what the CNN prediction is based on.

%% file: Introduction.tex
\label{sec_introduction}

Measuring poverty with remote sensing data has been a lively research field since the introduction of nighttime lights in the late 1990's \cite{elvidge_mapping_1997}. In a seminal study, Jean et al. \cite{XieEtAl2016} \cite{JeanEtAl2016} extended this work and showed how to fine-tune a deep Convolutional Neural Network (CNN), pre-trained on the ImageNet data, to predict poverty levels from high-resolution daytime satellite images in sub-Saharan Africa. This paper has been followed by many studies, see \cite{CastroA2022} for a review.
%
However, despite the successes of this approach, the results have yet to be employed for policy decisions. A recent review \cite{HallOR2022} showed that little effort has been devoted to explaining why CNNs make such good predictions of poverty levels, which may be a reason why there is hesitation to use these predictions in applications.

We show that high accuracy wealth estimates can be achieved  with lower resolution satellite images, with an accuracy higher than if human experts are asked to rate the wealth level from high-resolution satellite images. In addition, several approaches are explored to explain what features or details in the images that the CNNs base their predictions on.

%% file: Background.tex
\label{sec_Background}


Since 2016, there have been several applications of machine learning to poverty prediction. Many have used the transfer learning approach on high-resolution images and sometimes there have been attempts at interpreting the model, i.e. exploring what parts of the image that the model reacts to. Jean et al. \cite{XieEtAl2016} \cite{JeanEtAl2016} show that the CNN filters react to structures in the landscape, with more reactions in urban areas than in rural areas. An analysis of if and how these filters contribute to the Wealth Index predicted value is not provided. Liu et al. \cite{LiuEtAl2021} do gradient visualization with guided back-propagation in order to see what features give high night-time light intensity prediction. They conclude that for cities, these are contours of roads and areas of buildings are important. Tan et al. \cite{TanEtAl2020} use low-resolution satellite images and multitask learning for the transfer learning, i.e. more tasks than night-time lights. They show some examples of CNN filter activations for the fine-tuned network, but no analysis of how these filter activations relate to the wealth prediction. A general comment on the transfer learning features are that the illustrated filter responses may have been pretty much the same with a CNN trained on the ImageNet data, before fine tuning and without any connection to poverty.

Pandey et al. \cite{PandeyAK2018} 
train a CNN directly from high-resolution satellite images using multitask learning with labels from the 2011 Census of India. These tasks are, e.g., roof cladding material, source of drinking water, and source of lighting (a total of 24 such tasks). They illustrate, with some examples, that the CNN filters react to edges and seem to segment the images into farmland or human settlements, but there is no analysis of how these features relate to the Wealth Index prediction.

In short, published peer-reviewed work on explaining or interpreting CNN models' responses have been anecdotal, illustrating with a few filter responses but never analyzing how the filter responses relate to predicted Wealth Index values.

A different approach to using machine learning and high-resolution satellite images for wealth prediction have used features, or objects, extracted from the satellite images and then algorithms like random forest used to model the Wealth Index from the extracted features. Ayush et al. \cite{AyushEtAl2020} fine-tune a YOLOv3 model \cite{redmon2018yolov3} to detect objects in the satellite images. They model the Wealth Index with gradient boosting trees and apply SHAP analysis \cite{LundbergL2017} to the final model. Their conclusion is that trucks and other vehicles are the most important features; the number of trucks in the image being by far the most important feature. Huang et al. \cite{HuangHG2021} train a roof detector and claim that changes in wealth can be detected through changes in roof material.
Zhao et al. \cite{ZhaoEtAl2019} feed several features, computed from satellite images and other sources, into a random forest model and conclude that distance to closest urban area and night-time light radiance are the two most important features for wealth prediction.
Thus, the conclusions from the feature based wealth prediction vary and depend on the researchers' decision on which features to feed into the model. The feature based studies have not provided consistent clues on what the direct CNN-based Wealth Index prediction is based on.

Tang et al. \cite{TangEtAl2022} use the low-resolution normalized difference vegetation index (NDVI) as input. The CNN model is trained using the same transfer learning trick as used by Jean et al. \cite{JeanEtAl2016}. They then use random forest for estimating the Wealth Index from the CNN features, concluding that NDVI-based predictions are as good as the predictions achieved with daytime satellite images. This shows that color is important but not what structure or other features that are important.

Finally, the current state-of-the-art on predicting wealth from satellite images appears to be the recent work by Lee and Braithwaite \cite{LeeB2022}, who use both high-resolution daytime satellite images as well as OpenStreetMap data. Their approach is to iterate between a feature based prediction and the satellite image based prediction and thus achieve improved predictions. We have not used their model as the starting point for our study since their code is not available and there are many details in their approach that are hard to replicate. For example, they compute their own International Wealth Index, they manually correct survey site locations, and they compute several complicated features from OpenStreetMap data.

%% file: Data.tex
\label{sect_data}

Our study is based on three data sources: satellite data, both day-time imagery and night-time data, and survey data on household status. All data are for countries in sub-Saharan Africa.

\subsection{Household survey data}
The Demographic and Health Surveys (DHS) program is a key source of data for researchers and policymakers interested in understanding health, population, and social dynamics in low- and middle-income countries. DHS surveys collect nationally representative data on a wide range of health and social indicators, including assets wealth. DHS data are publicly available and can be accessed through the DHS program website upon request.

The surveys are cross-sectional, they cannot establish causality or track individual-level changes over time. The methodology is however standardized, ensuring that data is collected using consistent and rigorous procedures across all countries, making it possible to compare data across countries and over time, and thus facilitating cross-country analyses and comparisons. Surveys are conducted 3-5 years apart, sometimes with much longer intervals. Survey waves are uncoordinated between countries resulting in temporarily heterogeneous data. To protect the privacy and confidentiality of survey participants, DHS surveys scramble the GPS coordinates of the households and clusters by a small, random amount. This process, known as "geographic displacement", ensures that the precise location of the households and clusters are anonymized.

The larger dataset contains a total of 64,864 clusters across 34 Sub-Saharan Africa countries, spanning the period between 1990 and 2019. A subset of this dataset was drawn from the 2015-16 Tanzania DHS dataset which covers 12,563 households, aggregated into 608 clusters across the 30 administrative regions of the country. Our human evaluators were tested on this subset data.

As Wealth Index value, we use the mean of the Wealth Index factor score combined, with five decimals
This is in line with previous work, e.g. Head et al. \cite{HeadEtAl2017}.

\subsection{Satellite images}

Three sets of satellite data were used: Sentinel-2 daytime images for the CNN network,  Visible Infrared Imaging Radiometer Suite (VIIRS) night-time images to get night-time light intensities, and high-resolution Google Maps images for humans to view.

Sentinel-2 and night-time light images were downloaded from Google Earth Engine in GEOTIFF format.
Night-time light images came from the
VIIRS Stray Light Corrected Nighttime Day/Night Band Composites Version 1, with a resolution of 750 meters per pixel, and
daytime images came from Sentinel-2 MSI: MultiSpectral Instrument, Level-1C, bands B2, B3, and B4, with a resolution of 10 meters per pixel.
Both daytime and night-time images were downloaded for the year 2016,
daytime images were retrieved with cloud filtering percentile p45.

Sentinel-2 daytime images were split up into sub-images of $224 \times 224$ pixels, to fit as input to the MobileNet-V2 network. This corresponds to areas of $2240 \times 2240$ square meters. The VIIRS night-time light images were split up into matching sub-images of $3 \times 3$ pixels, corresponding to $2250 \times 2250$ square meters. The summed light intensity over the $3 \times 3$ pixel sub-image was used as the night-time light value for the corresponding daytime satellite image. If the light intensity for a VIIRS pixel was below 0.5 then it was set to zero, based on an estimate of the noise level. All images with night-time light values $\geq 1$ and a random selection of images with night-time light values $< 1$ were used for the CNN training.

Three datasets of Sentinel-2 and VIIRS data were collected, with increasing sizes. The smallest, denoted ``TZA'', consists of images from Tanzania only. This data set has 325,785 images; of which 323,966 (99.4\%) have night-time light values $< 1$. The medium sized one, denoted ``TZA + 11 cities'', consists of images from Tanzania plus images from 11 cities in neighboring countries. This data set has 333,221 images; of which 330,198 (99.1\%) have night-time light values $< 1$. The largest data set, denoted ``TZA + surroundings'', consists of images from Tanzania plus images from neighboring countries. This data set has 2,030,657 images; of which 1,848,356 (91.0\%) have night-time light values $< 1$. The group of images with night-time light $< 1$ were heavily downsampled for all CNN trainings. Typically, for a CNN training session, 58\% of the images would have night-time light $< 1$.





The Google Maps Platform hosts a set of APIs through which developers can retrieve data from the platform. Images are from different sensors and combined into a mosaic of images taken over multiple periods. Some of the sensors used are Landsat 8, Pleiades 1A, Quickbird and WorldView 4. For the study with human evaluators, 608 images were downloaded at zoom level 18, which corresponds to a resolution of about 0.6 meters per pixel. These images covered 608 Tanzania DHS sites, but with the location corrected to adjust for the random displacement in the DHS data. The correction was based on a guess of the correct location. 





\subsection{Human evaluation data}

The main inclusion criteria for our human experts is to be development researchers, broadly defined, with some fieldwork experience in any region of SSA. The human evaluators were shown a batch of high-resolution satellite imagery corresponding to each DHS cluster (village or city) location in a sequential manner and asked to rate the level of welfare on a five-grade scale ranging from \emph{Poorest}, \emph{Poorer}, \emph{Middle}, \emph{Richer}, to \emph{Richest}; akin to the DHS wealth quintiles. The median rating for each cluster image was then taken as representing the relative poverty score for the cluster. Raters were allowed to rate additional batches of 30 images if they so wished. Overall, we received 2,174 ratings from 102 experts.  
As can be seen from Table 1, a significant majority of our human evaluators were males, with more than half of the sample being between 36 and 55 years old and a further 18 per cent being even older. In terms of educational attainment, half of the human evaluators have a Ph.D. as the highest education level attained. With regards to the main region of fieldwork experience in SSA, they are fairly well-distributed - 47 per cent for West Africa, 36 per cent for East Africa, including Tanzania, and 17 per cent for the Central and Southern Africa region. For the length of experience, the sample has substantial field experience in these regions, with exactly half of the sample having more than 10 years of fieldwork experience. A summary description of our human experts is presented in Table~\ref{tab:experts}.

\begin{table}[]
    \centering
    \begin{tabular}{|l| l |c|}
    \hline
        \multicolumn{2}{|c|}{Variable Description} &  Proportion(\%)\\ \hline
        \multirow{3}{*}{Sex}& Females & 22.5\\
        & Males & 75.5 \\ 
        & Non-binary & 2.0 \\ \hline
        \multirow{3}{*}{Age Distribution}& Up to 35 years old & 30.4\\
        & Between 36 and 55 years old & 52.0 \\
        & 56 years and older & 17.6 \\ \hline
        \multirow{3}{2.4cm}{Highest Educational Attainment}& Bachelor's Degree& 16.7\\
        & Master's Degree& 33.3\\
        & Ph.D. & 50.0 \\ \hline
        \multirow{3}{2.4cm}{Region of Experience} & East Africa, incl. Tanzania & 36.3 \\
        & Central and Southern Africa & 16.7 \\
        &West Africa & 47.1 \\ \hline
        \multirow{3}{2.4cm}{Length of Experience} & Between 1 and 10 years & 50 \\
        & Between 11 and 20 years & 30.4 \\
        & 21 years and above & 19.6 \\ \hline
    \end{tabular}
    \caption{Summary Description of Human Experts (N=102)}
    \label{tab:experts}
\end{table}

%% file: Methods.tex
\label{sect_methods}

\subsection{Modeling the Wealth Index}






MobileNet-V2 was taken as starting point for the model. It is a 53 layer deep CNN pre-trained on the ImageNet database. It was trained to model night-time lights in a two-stage process, first adjusting only the weights in the final layer to the output, and then fine-tuned by also adjusting weights earlier in the network. The first step was done for 20 epochs, and the fine-tune training was done for 20 epochs. The learning rate was set to $0.01$ in the first part, using Adagrad training, and then the learning rate was reduced to $0.001$ during fine-tuning, using L2 regularisation with parameter 0.1. The batch size was set to $100$.
The MobileNet-V2 takes the $224 \times 224$ daylight images as input and is trained to output the corresponding $\log (1 + \mbox{night-time light})$.

Once the MobileNet-V2 network is trained, the values from the last layer before output (1280 units) are used as features for a linear regression model of the Wealth Index. Ridge Regression is used to fit the linear regression, using a cross-validation approach to set the ridge parameter. The Ridge Regression is done several times with different training and validation data to gauge the stability of the solution. All predictions of Wealth Index are done on images that are not similar to images in the training and validation sets, and the predictions are repeated so that a prediction is done for each of the DHS survey sites.

The accuracies of the Wealth Index estimations are evaluated using three measures: the coefficient of determination $R^2$, the Spearman's rank correlation, and the Matthew's correlation coefficient when grouping into quintiles. 

The $R^2$ coefficient is standard in studies of wealth estimation and allows the results to be compared to earlier published results. Spearman's rank correlation is the correlation between two ranked estimates of the Wealth Index; it varies between $-1$ and $+1$, with $+1$ indicating perfect agreement between the two rankings. 
The quintile grouping is used specifically to compare against human estimations of wealth level, since it is probably easier and more reliable for a human to group places into wealth groups rather than to provide a full ranking of all the sites. The correctness of the groupings are measured using the Matthew's correlation coefficient (MCC) for each dichotomy of one group versus the others. 


\subsection{XAI methods}
\label{xai_methods}

Several methods were employed to explore what features the CNN based its prediction on. None of the XAI methods required any new training of the CNN network, but the ridge regression step was always repeated for modified images.

\subsubsection{Grid Shuffling}
The input image is divided into equal tiles using a square grid pattern, and then tiles are randomly shuffled in the image, breaking up structures that are larger than the tile sizes \cite{ZhangZ2019,crowder_robustness_2022}. Tile sizes range from one pixel to half the width of the image, and when tiles are $1 \times 1$ pixels all structural information in the image is lost after shuffling and only color information is left. The aim is to find the grid size when the Wealth Index prediction starts to deteriorate, indicating the size of structures in the image that the prediction is based on.

\subsubsection{Frequency Domain Filtering}
This consists of low-pass, high-pass, and band-pass filters. Filtering is done in the image frequency space and is done individually for each color in the image. A low-pass filter removes high frequencies and keeps the lower frequency information. A high-pass filter is the complement of the low-pass filter and keeps higher-frequency information. A band-pass filter keeps information between two frequency thresholds in the image. We use Gaussian filters throughout for frequency domain filtering, with a signal domain standard deviation equivalent of $\sigma$ for the low- and high-pass filters. The band-pass filter has a lower limit standard deviation of $\sigma$ and a higher limit standard deviation of $1.5\sigma$ in signal space. The frequency domain filtering affects the color of the image but should be able to confirm what scales of features in the images that are important for the prediction.


\subsubsection{L*a*b* Color Space Analysis} The L*a*b* color space is designed to correspond well to human perception \cite{robertson_cie_1977}. In this color space, the lightness (L* component) is decoupled from the chromaticity of the color (a* and b* components), which allows separating objects based on material, regardless on the lightness variations. We used the L*a*b color space to cluster the image pixels based on chromaticities, and segment out different types of materials in the image. By keeping the chromaticities of one type (cluster) of material and removing those of other types, we can compare how chromaticities (e.g. different materials) affect the Wealth Index estimation.

\subsubsection{Feature Attribution Methods}
These are methods that aim to compute the importance that each part of the image has for the CNN output. We used Occlusion Sensitivity \cite{ZeilerEtAl2014}, Grad-CAM~\cite{selvaraju_grad-cam_2020}, Guided Backpropagation~\cite{springenberg_striving_2015}, and Guided Grad-CAM~\cite{selvaraju_grad-cam_2020}. 

Occlusion Sensitivity shows the change in the output if a square patch in the image is grayed out. The change in output is calculated for locations on a grid and the result is visualized as a map that should show the parts of the image that are important for the output, and if they have a positive or a negative effect on the output. 

Grad-CAM is a gradient-based method that uses the last convolutional layer to show a high level map of the importance of image regions on the output. Because the last convolutional layer is used, the resolution of the importance map is low, but it contains high level (as opposed to pixel level) information regarding the CNN output. 

Guided backpropagation is also a gradient based method, but it estimates the importance of each pixel for the CNN output. It creates a high resolution map (the same size as the input image) with pixel-level information about the CNN output. Since Grad-CAM and guided backpropagation are complementary regarding the level of information they provide, they have been combined by multiplying the two maps with each other. This combination is denoted Guided Grad-CAM.

\subsubsection{Feature Visualization} This is a recent method designed to show what patterns in the image different neurons in the CNN react to \cite{olah_feature_2017}. The method starts from an image with randomly assigned pixel values and then the image is optimized to maximize the output of a particular selected neuron in the CNN. Since the optimization problem certainly has many local maxima, the visualization will show an image that maximizes the output locally. The resulting \emph{optimal} image can provide hints on the type of image features that specific neurons respond to, but it is not guaranteed that the method produces a meaningful and interpretable visualization.

%% file: Results.tex
\subsection{Wealth Index estimation}

The first study was done to check if the use of low-resolution satellite images and the transfer learning approach would yield Wealth Index predictions that were as accurate as those with high-resolution images.

\begin{figure}[!t]
\centering
\includegraphics[width=2.5in]{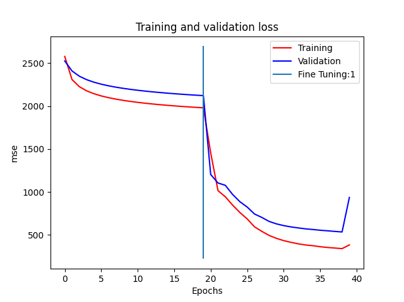}
\caption{Training and validation loss during the training of the CNN for night-time light prediction. The vertical bar indicates the start of the fine-tuning.}
\label{fig_training}
\end{figure}

Figure~\ref{fig_training} shows the training and validation errors during the CNN training on night-time lights values. It shows clearly that the CNN learns to predict the night-time lights and that the fine-tuning step is important. Table~\ref{table_firstres} summarizes the Wealth Index prediction accuracies when the 1280 CNN features are used as input in the ridge regression model, for three different training set sizes for the night-time lights estimation and two ways to produce the features used in the Ridge Regression. The three different training sets are ``TZA'', ``TZA + 11 cities'', and ``TZA + surroundings'' and correspond to increasing number of images (see Section \ref{sect_data}). The two ways for producing the features are denoted ``$1 \times 1$'' and ``$3 \times 3$'', and correspond to using the features from one single input image of size $224 \times 224$, or the averaged features from nine images in a $3 \times 3$ grid centered on the DHS location (see Section \ref{sect_methods}). Note that the Wealth Index estimation itself is always done on the 608 DHS survey sites in Tanzania, regardless of the size of the training data used for the night-time light estimation. 

Table~\ref{table_firstres} shows that increasing data set sizes lead to improved estimates of the Wealth Index, although more so for the $R^2$ coefficient than for the rank correlation. Also, combining more images yields better estimates than using a single image. All this is in agreement with previously published findings.

Jean et al. report an $R^2$ value of 0.57 when using the CNN transfer method to model the Wealth Index for Tanzania (2010 DHS survey data) \cite{JeanEtAl2016}. In 2017, Head et al. replicated some of their experiments, albeit not for Tanzania, and concluded that they got pretty much the same values \cite{HeadEtAl2017}, with variations in the second decimal. Thus, the results in Table~\ref{table_firstres} are on par with results using higher resolution images.

\begin{table}[t!]
\renewcommand{\arraystretch}{1.3}
\caption{TZA Wealth Index Estimation From Low Resolution Images On the 608 DHS Sites}
\label{table_firstres}
\centering
\begin{tabular}{|l|c|c|c|}
\hline
 & \multicolumn{3}{c|}{Training data} \\
 \hline
 & TZA & TZA + 11 cities & TZA + surroundings \\
\hline
$1 \times 1$ ($R^2$) & 0.57 & 0.60 & 0.61 \\
$3\times 3$ ($R^2$) & 0.65 & 0.66 & 0.67 \\
\hline
$1 \times 1$ (rank corr.) & 0.70 & 0.72 & 0.73 \\
$3 \times 3$ (rank corr.) & 0.76 & 0.76 & 0.77 \\
\hline
\end{tabular}
\end{table}

The second study was done to gauge whether it mattered if survey data came from the same period as the satellite image data, or from times before or after. Here, the CNN network was always the same and the only thing that changed was the DHS data used in the ridge regression. The large DHS data set was split up into three parts: Tanzania survey results with Wealth Index for DHS survey phases 4 and 5, which meant years 2003 to 2008 (809 DHS sites), Tanzania survey results with Wealth Index for DHS survey phase 6, which meant years 2009 to 2012 (1,028 DHS sites), and Tanzania survey results with Wealth Index values for DHS survey phase 7, which meant years 2015 to 2017 (1,044 DHS sites). The three data subsets were used either as test or training sets (or a $k$-fold cross-validation if the time periods were the same). Table~\ref{table_crossres} shows the results of this experiment, always using $3 \times 3$ images, with $R^2$ values in the upper half and rank correlations in the lower half. The results are better for the later period, which agrees with the time for the satellite images, but the differences are small. This indicates that the features used in the prediction do not change very much over a time period of 5-20 years, or that the Wealth Index values do not change much relative to each other over this time period. 

\begin{table}[t!]
\renewcommand{\arraystretch}{1.3}
\caption{TZA Wealth Index Estimation Across Time}
\label{table_crossres}
\centering
\begin{tabular}{|l|c|c|c|}
\hline
 & \multicolumn{3}{c|}{DHS Test data}\\
 \hline
DHS Training data & 2003-2008 & 2009-2012  & 2015-2017 \\
\hline
2003-2008 ($R^2$) & 0.69 & 0.69 & 0.69 \\
2009-2012 ($R^2$) & 0.70 & 0.70 & 0.71 \\
2015-2017 ($R^2$) & 0.67 & 0.70 & 0.73 \\
\hline
2003-2008 (rank corr.) & 0.74 & 0.75 & 0.78 \\
2009-2012 (rank corr.) & 0.75 & 0.76 & 0.78 \\
2015-2017 (rank corr.) & 0.75 & 0.77 & 0.79 \\
\hline
\end{tabular}
\end{table}

The third study compared the prediction accuracy of the low-resolution CNN model, combining $3 \times 3$ low-resolution images, to that of human evaluators that viewed high-resolution images. This study was done on corrected DHS data, meaning that the location of the DHS survey data had been adjusted to the position believed to be the correct DHS survey position for the cluster. The CNN Wealth Index predictions where grouped into quintiles, from poorest to wealthiest. The human subjects were similarly asked to group the site they saw into one of five categories, from poorest to wealthiest. 

The two confusion matrices below show how the Wealth Index predictions from the CNN and from the human evaluators compare to the quintile grouping based on the actual DHS survey result. Group 1 corresponds to the 20\% poorest, and group 5 corresponds to the 20\% wealthiest. An observation is that the human subjects label many more sites to be average, and fewer sites as very poor or very wealthy. Furthermore, in the average group 3, the human evaluators placed sites from all five wealth levels, almost with equal probabilities. This indicates that it is difficult for a human to estimate wealth level from a satellite image.

\[
\renewcommand\arraystretch{1.3}
\begin{array}{|c|c|c|c|c|c|c|}
\hline
    & & \multicolumn{5}{c|}{\mbox{Predicted group (CNN)}} \\
    \hline
    & & \bf{1} & \bf{2} & \bf{3} & \bf{4} & \bf{5} \\
    \hline
    \multirow{5}{*}{\STAB{\rotatebox[origin=c]{90}{\mbox{True group}}}} 
    & \bf{1} & 63 & 41 & 11 & 7 & 0 \\
    \cline{2-7}
    & \bf{2} & 25 & 44 & 29 & 13 & 0 \\
    \cline{2-7}
    & \bf{3} & 18 & 27 & 42 & 29 & 6 \\
    \cline{2-7} 
    & \bf{4} & 5 & 8 & 28 & 51 & 29 \\
    \cline{2-7}
    & \bf{5} & 1 & 1 & 12 & 21 & 87 \\
    \hline
\end{array}
\]

\[
\renewcommand\arraystretch{1.3}
\begin{array}{|c|c|c|c|c|c|c|}
\hline
    & & \multicolumn{5}{c|}{\mbox{Predicted group (hum.)}} \\
    \hline
    & & \bf{1} & \bf{2} & \bf{3} & \bf{4} & \bf{5} \\
    \hline
    \multirow{5}{*}{\STAB{\rotatebox[origin=c]{90}{\mbox{True group}}}} 
    & \bf{1} & 12 & 64 & 36 & 9 & 1 \\
    \cline{2-7}
    & \bf{2} & 13 & 54 & 47 & 5 & 2 \\
    \cline{2-7}
    & \bf{3} & 10 & 54 & 40 & 17 & 1 \\
    \cline{2-7} 
    & \bf{4} & 6 & 37 & 53 & 25 & 0 \\
    \cline{2-7}
    & \bf{5} & 4 & 24 & 49 & 40 & 5 \\
    \hline
\end{array}
\]

The correctness of the CNN and human groupings against the true DHS values were compared using four dichotomies in increasing wealth level: (1) whether a site was grouped into group 1, instead of any of groups 2-5, (2) whether a site was grouped into group 1 or 2, instead of any of groups 3-5, (3) whether a site was grouped into any of groups 1-3, instead of groups 4-5, or (4) whether a site was grouped into groups 1-4, instead of group 5. For each dichotomy, the Matthew's correlation coefficient (MCC) was computed.  

Table~\ref{table_mcc_quintile} shows the MCC values for the two approaches. The CNN predicted wealth levels are consistently better than the human estimates from high-resolution images. Thus, the CNN Wealth Index predictor is producing something quite impressive from the low-resolution images and there should be a significant value in trying to understand the predictions.

Notably, the CNN model had less accuracy on the wealth indices when using corrected DHS positions instead of the original DHS positions, which puts into question if the new positions were more correct than the original positions.

\begin{table}[t!]
\renewcommand{\arraystretch}{1.3}
\caption{Human and CNN Quintile Groupings of TZA Wealth Index}
\label{table_mcc_quintile}
\centering
\begin{tabular}{|l|c|c|c|c|}
\hline
\bf{MCC vals.} & 1 vs. 2-5 & 1-2 vs. 3-5 & 1-3 vs. 4-5 & 1-4 vs. 5 \\
 \hline
CNN & 0.17 & 0.31 & 0.38& 0.36 \\
\hline
Human & 0.03 & 0.09 & 0.13 & 0.16 \\
\hline
\end{tabular}
\end{table}

\subsection{Explorations of explanations}

Here, we present results with the XAI methods described in Section \ref{xai_methods}. For practical reasons, when results are reported for a $3 \times 3$ input it means that the nine images have been concatenated into a $672 \times 672$ pixel image and fed into a fully-convolutional network created by fusing nine CNNs. This is different from previous results where predictions were done using feature values averaged over nine CNNs. The XAI work was done on the CNN trained on the TZA dataset, cf. Table~\ref{table_firstres}.



\subsubsection{Grid Shuffling}

\begin{figure*}
    \centering
    \includegraphics[width=0.35\textwidth]{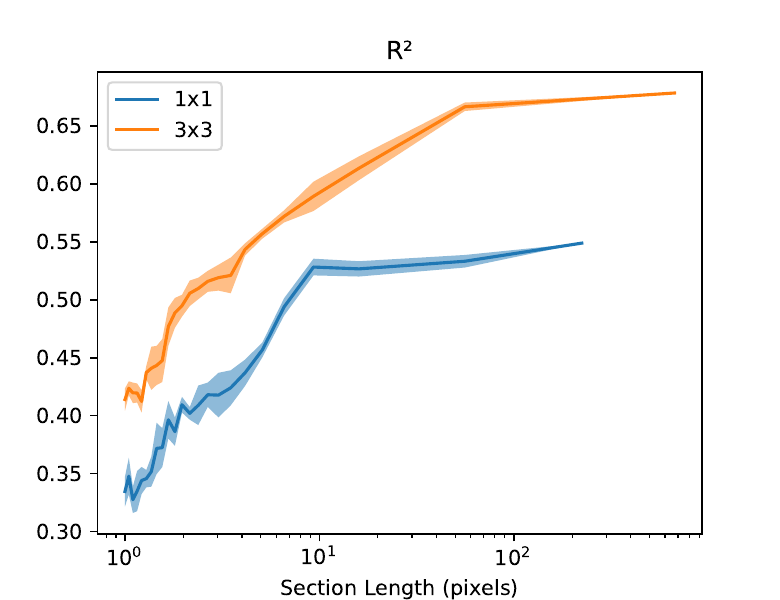}
    \includegraphics[width=0.35\textwidth]{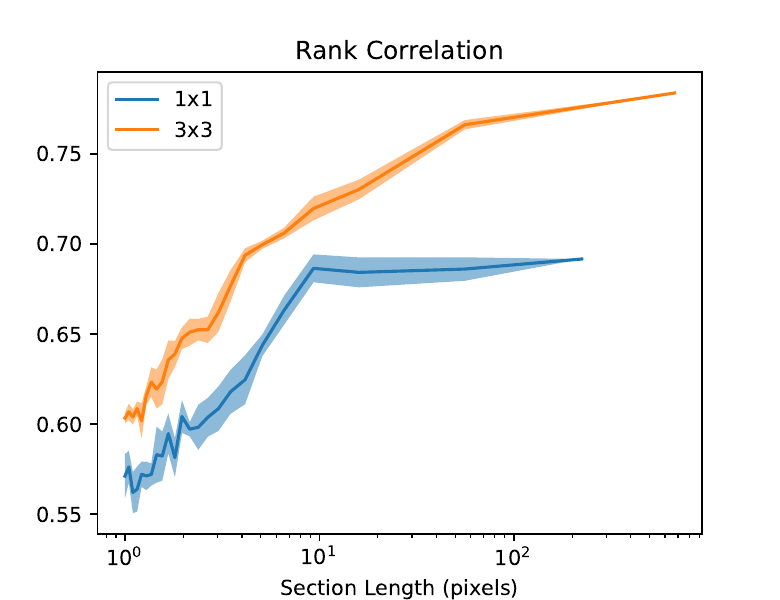}
    \caption{$\textrm{R}^2$ and Spearman's Rank Correlation performance for Wealth Index prediction when input images are split up in tiles and tiles are shuffled. Results for $3\times3$ inputs are shown in orange and results for $1\times1$ inputs are shown in blue. Five random shuffling experiments were performed for each tile size, and the curves show the mean and standard deviation.}
    \label{fig:shuffling}
\end{figure*}

The results from doing Grid Shuffling are shown in Figure \ref{fig:shuffling}. 
Wealth Index prediction performance is shown versus tile size (section length). The performance curve for the $3 \times 3$ input is consistently above the curve for the $1 \times 1$ input. This is expected since the results with averaged image features (i.e. combining images) are better in general. It is, however, surprising that the curves begin decreasing at different scales; $1 \times 1$ starts decreasing at 10 pixels, whereas $3 \times 3$ starts decreasing at approximately 60 pixels.


\subsubsection{Frequency Domain Filtering}

\begin{figure*}
    \centering
    \includegraphics[width=0.3\textwidth]{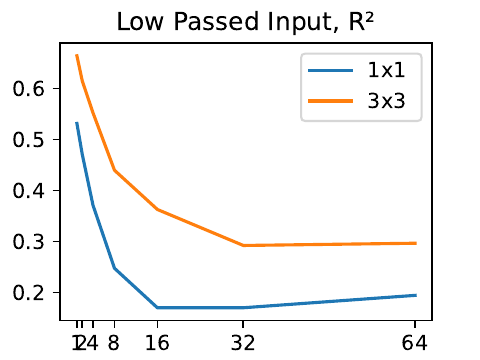}
    \includegraphics[width=0.3\textwidth]{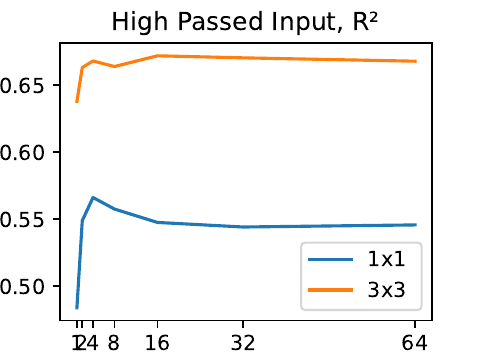}
    \includegraphics[width=0.3\textwidth]{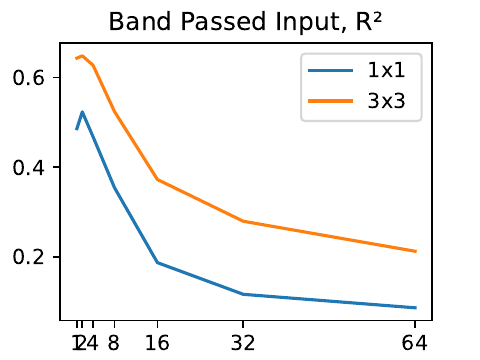}
    \caption{Wealth Index prediction performance of the CNN when provided with low-pass, high-pass, and band-pass filtered images as input. The x-axis indicates the standard deviation of the Gaussian filter, expressed in pixels. Only results for $R^2$ are shown to save space.}
    \label{fig:filters}
\end{figure*}

The effects of feeding low-pass, high-pass, or band-pass filtered images to the CNN are shown in Figure~\ref{fig:filters}. The x-axis label is 
%
\[
\sigma = \frac{N}{2 \pi D}
\]
where $D$ is the standard deviation of the Gaussian filter in frequency space and $N = 224$.

The low-pass filter leads to a general decrease in performance as $\sigma$ increases. It is only for $\sigma \leq 2$ that the prediction performance is acceptable, which means almost no filtering. The minimum performance for low-pass filtered images is considerably lower than for the Grid Shuffling, which is surprising but could be related to how color is affected by the filtering (the colors become more and more gray). Similarly, the high-pass filter appears to not affect the Wealth Index prediction very much before it comes to $\sigma \leq 2$, which corresponds to keeping only the highest frequencies in the image, i.e. objects 1-2 pixels in size, and discarding the rest. Together, the high-pass and low-pass results indicate that almost all scales in the image are important, but high frequency components are much more important than low frequency components.

In the band-pass filtered images details at a certain frequency (size) are kept in the image while others are grayed out. The Wealth Index prediction performance with band-pass filtered images maximizes around 2-3 pixels, and the performance is almost as good as with unfiltered images. This indicates that objects of sizes around 2-3 pixels (20-30 meters) are most important.

\subsubsection{L*a*b* Color Space Analysis}
The colors for each pixel for each one of the 608 DHS Sentinel-2 images were converted from RGB into L*a*b* space. Then $k$-means clustering was applied on the pixel distribution, with the intention to find meaningful concentrations of pixels in the L*a*b* space. The number of clusters ($k$) was set to 3 using the elbow method.

To measure the importance of each color cluster, the colors for pixels in each cluster were then modified in two different ways, one where the color of pixels in two clusters was removed but the lightness kept intact, and one where both color and lightness in two clusters were removed. In the first type of experiment, the chromaticities of pixels in two clusters were removed by setting their a* and b* components to zero, and the modified image was then fed to the CNN. In the second type of experiment, the pixels for two clusters were ``blanked out'' by setting them to an average gray color (setting the R, G, and B components to the middle of their valid range).

\begin{figure}
    \centering
    \includegraphics[width=0.48\textwidth]{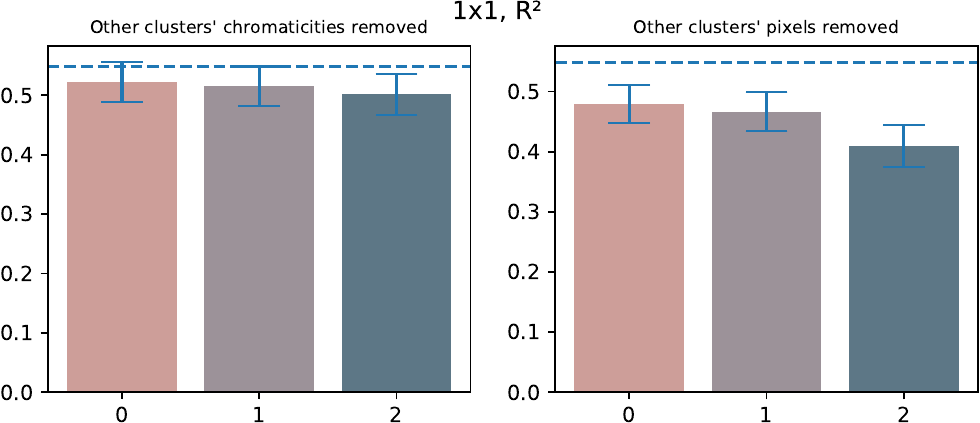}

    \includegraphics[width=0.48\textwidth]{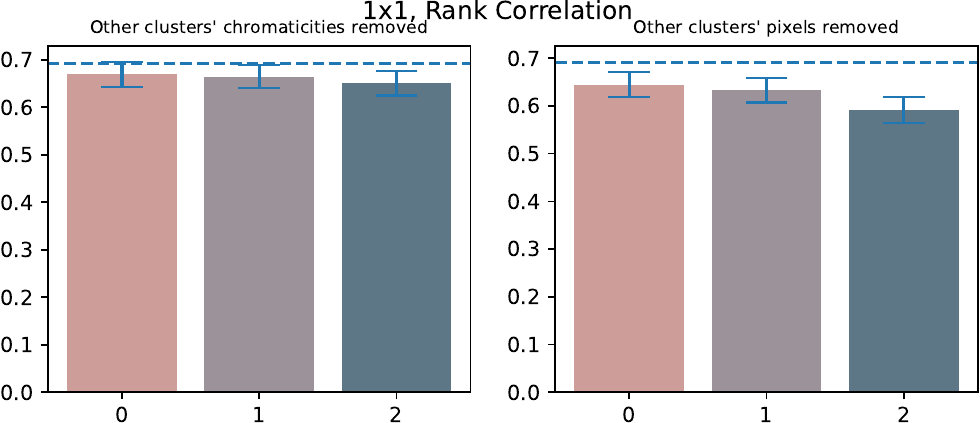}
    \caption{Performance bar plots for experiments on the L*a*b* space pixel clusters on $1 \times 1$ images. Two types of experiments were done for each cluster: first, removing the chromaticity of other clusters' pixels, and second, graying the pixel values of other clusters' pixels. The height is the average performance and the error bars are the standard deviation, estimated using bootstrapping. The horizontal dashed line corresponds to the CNN original performance.}
    \label{fig:color_modification_1x1}
\end{figure}

\begin{figure}

    \centering
    \includegraphics[width=0.48\textwidth]{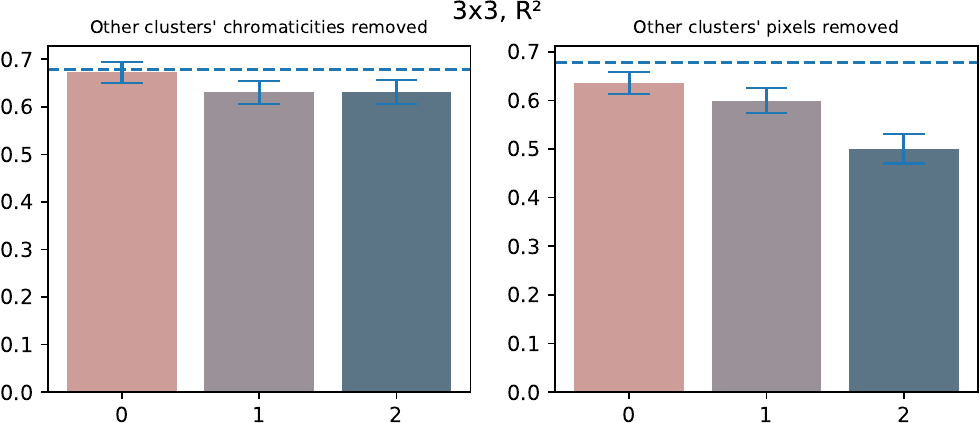}

    \includegraphics[width=0.48\textwidth]{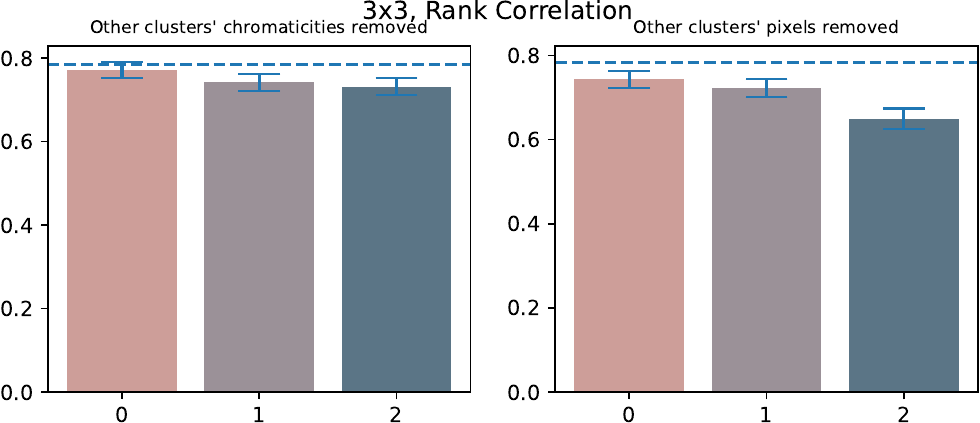}
    \caption{Performance bar plots for experiments on the L*a*b* space pixel clusters on $3 \times 3$ images. Two types of experiments were done for each cluster: first, removing the chromaticity of other clusters' pixels, and second, graying the pixel values of other clusters' pixels. The height is the average performance and the error bars are the standard deviation, estimated using bootstrapping. The horizontal dashed line corresponds to the CNN original performance.}
    \label{fig:color_modification_3x3}
\end{figure}

The results for these experiments are summarized in Figures~\ref{fig:color_modification_1x1} and \ref{fig:color_modification_3x3}. The colors of the bars are the cluster centers in L*a*b* space. One cluster is centered on dark orange/light brown color that could be related to dry areas. Another cluster is centered on a dark brown color that could be related to mountainous areas, buildings, or other types of infrastructure. The third cluster is centered on a bluish green color that could be related to green areas and bodies of water. Keeping the dark orange/light brown cluster intact but removing color for the two other clusters changes the performance very little. Graying the pixels completely always has a larger effect than just removing color, but the effect is surprisingly small.

\subsubsection{Feature Attribution Methods}

\begin{figure*}
    \centering
    \hspace*{-0.8in}\includegraphics[width=1.2\textwidth]{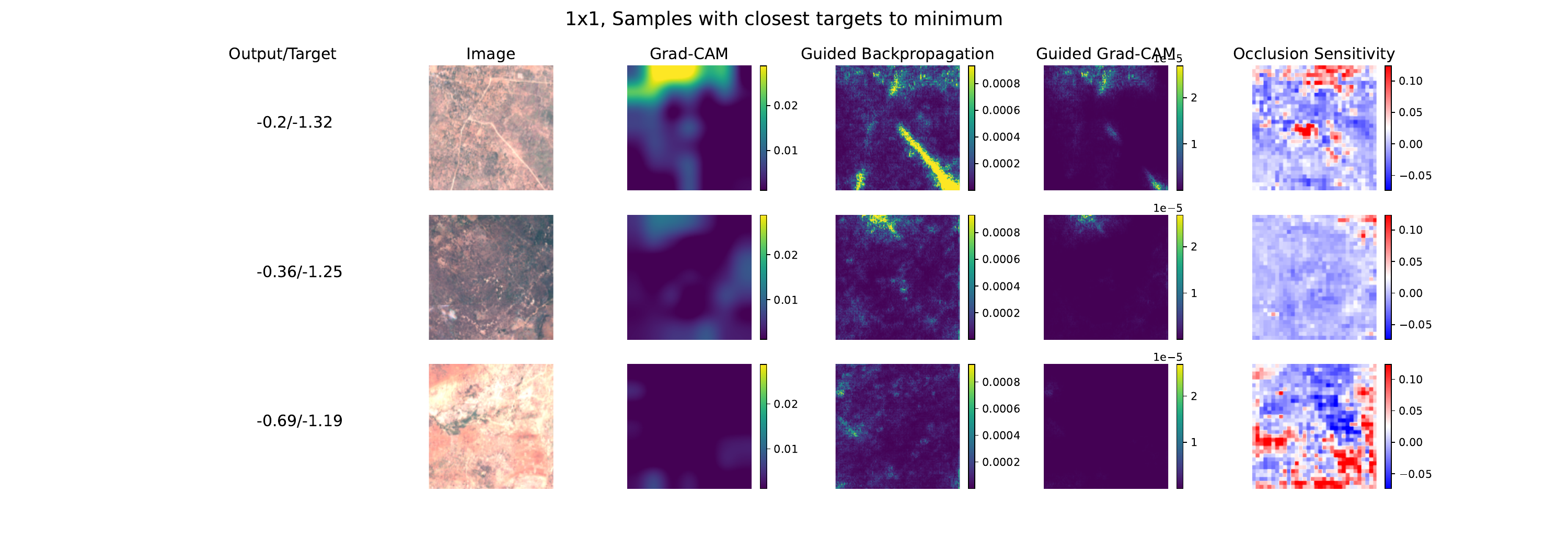}
    \hspace*{-0.8in}\includegraphics[width=1.2\textwidth]{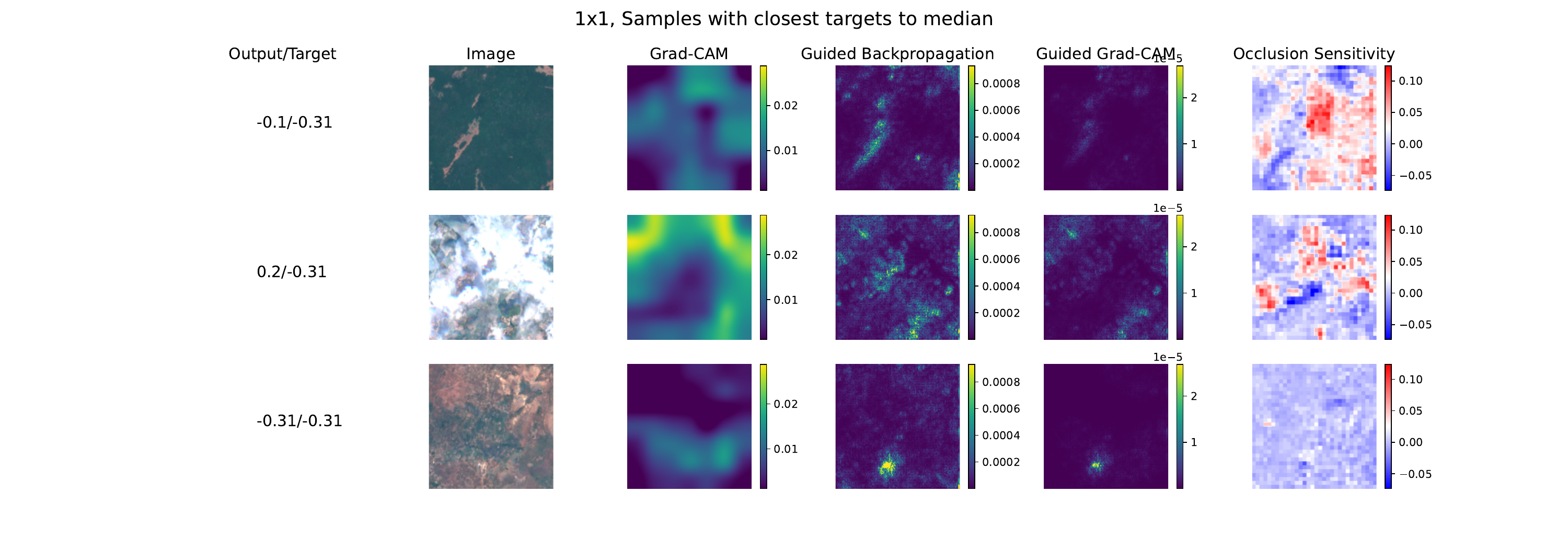}
    \hspace*{-0.8in}\includegraphics[width=1.2\textwidth]{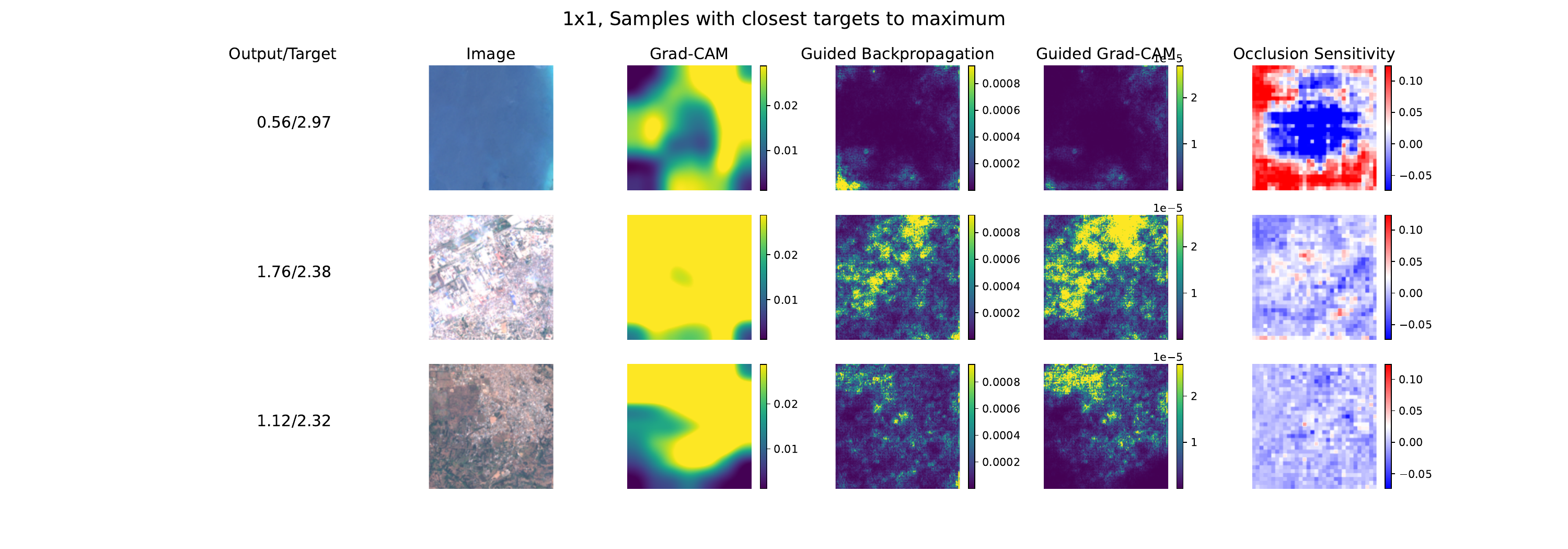}
    \caption{Feature attribution methods for $1\times1$ input samples. From the testing set, three samples with the lowest target values, three samples with highest target values, and three samples with target values closes to the median are used for visualizations.}
    \label{fig:1x1_NN_explainability}
\end{figure*}

Figure~\ref{fig:1x1_NN_explainability} shows the results from applying Grad-CAM, Guided Backpropagation, Guided Grad-CAM, and Occlusion Sensitivity to some sample $1 \times 1$ images in the 608 DHS data set (each image is centered on a DHS survey location).
Grad-CAM has higher outputs for samples with larger Wealth Index values but because of the low resolution it is not very helpful in giving detailed information on how the network is making the decision. It sometimes lights up in places where there are buildings or man-made infrastructure, but it is not consistent. Furthermore, in one case where the image is in the ocean Grad-CAM highlights several regions, which is hard to interpret (the image shows the ocean due to the random displacement applied to the true DHS survey location). Of the remaining Feature Attribution Methods, Guided Grad-CAM seems to be the most consistent in detecting human infrastructure and not reacting to blue ocean. However, none of the methods provide a very convincing basis for explaining the output value.

A test was made to see how well the features from Grad-CAM, Guided Backpropagation, and Guided Grad-CAM matched with the CNN output. The sum of the responses in the images was computed and correlated with the CNN output value. The correlations are shown in Table~\ref{table_feature_sum_CNNout}. The correlation between Grad-CAM and CNN output is very high, but Grad-CAM is computed for the last convolutional layer so this should be expected. However, the correlation between the sum of Guided Grad-CAM and the CNN output is also high, higher than for Guided Backpropagation, which indicates that the features highlighted for Guided Grad-CAM explain a lot.

\begin{table}[t!]
\renewcommand{\arraystretch}{1.3}
\caption{Correlation Between CNN Output and Summed Feature Attributes}
\label{table_feature_sum_CNNout}
\centering
\begin{tabular}{|l|c|c|c|}
\hline
\bf{Correlation} & Grad-CAM & Guided BP & Guided Grad-CAM \\
 \hline
CNN output & 0.99 & 0.79 & 0.89 \\
\hline
\end{tabular}
\end{table}

\subsubsection{Feature Visualization}
\begin{figure*}
    \centering
    \includegraphics[width=0.25\textwidth]{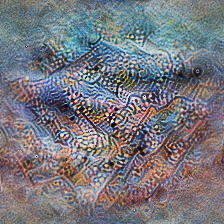}
    \includegraphics[width=0.25\textwidth]{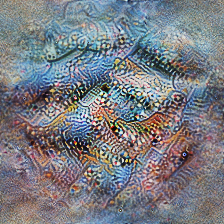}
    \includegraphics[width=0.25\textwidth]{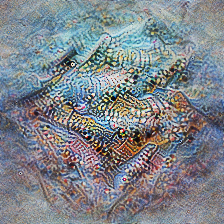}
    
    \caption{Feature visualization of the most optimal input image based on the output neuron for our Wealth Index regression neural network. The three examples were created starting from three different random initial images.}
    \label{fig:feature_visualization}
\end{figure*}

The result from applying the Feature Visualization technique on the output neuron that predicts the Wealth Index is shown in Figure~\ref{fig:feature_visualization}.
The three visualizations correspond to three different initial random images. This shows what type of input image that would maximize the Wealth Index output, and it is not hard to imagine that it looks like a populated place with roads, streets, houses and neighborhoods.


%% file: Discussion.tex

This study was initiated to understand what it is in daytime satellite images that a CNN reacts to and enables such good predictions of poverty status, better than estimates made by a human evaluator from high-resolution satellite images. Previous analyses of, e.g., filter responses in the CNN have been anecdotal, providing a few example images but never considering how the features affect the complete set of predictions.

The studies with Grid Shuffling indicate that details 100-600 meters in size are important. The Gaussian frequency filter experiments indicate that small details, 20-30 meters in size, are important and that large details (low frequencies) are not informative. Low-pass filtering the image destroys the Wealth Index prediction. Color only is not sufficient, since shuffling the individual pixels retains color but the Wealth Index prediction is quite poor.

Both the Grid Shuffling and the Gaussian filtering experiments show that combining $3 \times 3$ images instead of just the single $1 \times 1$ images is always beneficial. This was not immediately obvious since one would think that objects of a certain size should be useful in wealth index estimation regardless of the input image size. However, there are at least a couple of reasons why combining images would be helpful. First, due to the (small) random displacement of the correct DHS locations, using neighboring images to the stated DHS location increases the probability of one of the images covering the correct survey site. The second relates to the statistical probability for covering important information. If the Wealth Index prediction is based on the number of small infrastructure features found in an image, then having more images increases the probability for detecting enough of these small features.

The L*a*b* color experiments show that the Wealth Index prediction can be done well even if the color and intensity of many pixels are set to an average gray, and just one color cluster is kept. This result is in line with earlier results by Tang et al.~\cite{TangEtAl2022} who showed that it is possible to predict the Wealth Index with quite high accuracy ($R^2 = 0.60$ for Tanzania) using low-resolution satellite images with the normalized difference vegetation index (which probably is close to the complement of the orange/brown color we get in the L*a*b* clustering). However, the L*a*b* experiments do not support the supposition that roof cladding material is a very important feature. Also, the resolution of the images (10 meters per pixel) probably makes this feature difficult to detect.

At first, the Feature Attribution results seemed not very informative or consistent across samples, but the correlation between the sum of the feature values and the CNN prediction indicate that they describe a lot of the CNN decision. The Grad-CAM, which has the highest correlation with the CNN ouput, has too low resolution to be consistent with our other findings. However, the Guided Grad-CAM looks very promising as a basis for explaining the CNN response. The features it highlights match other findings about size, and it looks like it is basically the number of these features in the image that determines the CNN output.

Finally, the image Feature Visualization of the output neuron gives indications about what an image with a high Wealth Index looks like. The visualizations appear like the street structure of a city and buildings inside it. Main streets and minor streets are shown as bigger and straight, or smaller and curvy lines, respectively. Building structures can be seen as dots. It should be possible to further investigate Feature Visualization by systematically visualizing the features from layers before the output layer.

%% file: Conclusion.tex

The experiments indicate that a CNN that predicts Wealth Index for Tanzania based on low-resolution satellite images forms its prediction from many small details in the image; roads, road networks, and buildings. How many such small details there are in the image appears important, but the exact material in buildings appears to not be important, and neither is the number of vehicles. The resolution in the images used is too low to describe individual vehicles but the Wealth Index prediction is still competitive with that using much higher resolution images, and apparently better than predictions made by human evaluators using high-resolution images.